\documentclass{article}
\usepackage{spconf}

\usepackage[utf8]{inputenc} 
\usepackage[T1]{fontenc}    
\usepackage{hyperref}       
\usepackage{url}            
\usepackage{booktabs}       
\usepackage{amsfonts}       
\usepackage{nicefrac}       
\usepackage{microtype}      
\usepackage{xcolor}         
\usepackage{amsmath}
\usepackage{amssymb}
\DeclareMathAlphabet\mathbfcal{OMS}{cmsy}{b}{n}
\usepackage{amsthm}
\usepackage{xurl}
\usepackage{placeins}
\usepackage{bm}
\usepackage{svg}
\usepackage{graphicx}
\usepackage{tabularx}
\usepackage{comment}
\usepackage{multirow}
\usepackage{adjustbox}
\usepackage{subcaption}

\usepackage{siunitx}

\graphicspath{{./images/}}

\newcolumntype{T}[1]{S[table-format=#1]}

\title{Time Changed Normalizing Flows for accurate SDE modeling}
\name{Naoufal El Bekri$^{1,2}$, Lucas Drumetz$^{2}$, and Franck Vermet$^{1}$\thanks{This work was supported by Agence Nationale de la Recherche under grant
ANR-21-CE48-0005 LEMONADE, and by France 2030 framework program, Centre Henri Lebesgue, under grant No ANR-11-LABX-0020-01.}}
\address{$^{1}$Univ Brest,  UMR CNRS 6205, Laboratoire de Mathématiques de Bretagne Atlantique, France \\
$^2$IMT Atlantique, Lab-STICC, UMR CNRS 6285, Brest, France }

\begin{document}

\maketitle

\begin{abstract}
The generative paradigm has become increasingly important in machine learning and deep learning models. Among popular generative models are normalizing flows, which enable exact likelihood estimation by transforming a base distribution through diffeomorphic transformations. Extending the normalizing flow framework to handle time-indexed flows provided dynamic normalizing flows, a powerful tool to model time series, stochastic processes, and neural stochastic differential equations (SDEs). In this work, we propose a novel variant of dynamic normalizing flows, a Time-Changed Normalizing Flow (TCNF), based on time deformation of a Brownian motion which constitutes a versatile and extensive family of Gaussian processes. This approach enables us to effectively model some SDEs that cannot be modeled otherwise, including standard ones such as the well-known Ornstein-Uhlenbeck process, generalizes prior methodologies, and leads to improved results and better inference and prediction capability.
\end{abstract}


\section{Introduction}
Dynamical systems have widespread use in various scientific areas such as finance, geosciences, and physics. The representation of these systems usually involves Ordinary Differential Equations (ODEs), or Stochastic Differential Equations (SDEs)~\cite{oksendal2013stochastic} when noise and perturbations, on top of the deterministic component, are considered. Crucial applications include modeling volatility in financial data, or uncertainty quantification and propagation in geosciences. Tackling such systems through time series modeling and machine learning is an approach that experienced a surge in popularity, particularly recently, thanks to the generative paradigm, for forecasting applications, filtering, or interpolation with a notion of uncertainty in the generated sequences.

Popular generative models include Generative Adversarial Networks (GANs)~\cite{conf/nips/GoodfellowPMXWOCB14} and variational autoencoders (VAEs)~\cite{journals/corr/KingmaW13}, but also more recently normalizing flows (NFs)~\cite{jmlr/PapamakariosNRM21, journals/pami/KobyzevPB21} and diffusion/score based models~\cite{conf/nips/SongE19}. Though these models can be applied to generate time series, they are not well-suited for the task because they treat such data as vectors in $\mathbb{R}^{T}$, with $T$ the number of time steps, without accounting for the causal structure. Adaptations of GANs, VAEs, and NFs to time series data have been carried out in~\cite{conf/nips/YoonJS19, kidger2021neural},\cite{li2020scalable, zeng2023latent},\cite{mehrasa2019point, conf/iclr/ShchurBG20}, respectively. In this work, we focus on NFs for their capacity to access explicit likelihoods, which is crucial for applications when uncertainty quantification or anomaly detection is required.

NFs are based on the celebrated change of variables formula that provides an expression of the probability density function of diffeomorphic transformations of a random variable. With carefully chosen transformations (or compositions thereof), if the initial density is tractable (explicit likelihood and easy sampling, in most cases Gaussian), the transformed density can be easily manipulated and sampled as well, provided that the Jacobian of the transformation can be computed efficiently. By considering the theoretical limit where an infinite number of transformations is applied, we can derive the Continuous Normalizing Flow (CNF)~\cite{DBLP:conf/iclr/GrathwohlCBSD19}. In this instance, the NF is described by an ODE that can be integrated to obtain the resulting density. This approach further improves the computational efficiency of this class of models by replacing the Jacobian determinant with the integration of its trace.

NFs have been extended to the dynamic setting by replacing the tractable base distribution with a tractable \emph{stochastic process}, i.e. a Brownian motion~\cite{deng2020modeling}, making this type of model much more efficient for time series generation. However, it was noted in~\cite{deng2021continuous} that these models are theoretically unable to handle some of the most basic and common processes, such as the classical Ornstein-Uhlenbeck process.

Thus, in this paper, we propose a generalization of these approaches by using a large family of Gaussian processes as the underlying base process instead of the conventional Brownian motion. The Gaussian processes are constructed by transforming through time the standard Brownian motion, giving rise to Time Changed Normalizing Flow (TCNF), a model that possesses mathematical properties that enable it to describe dynamics and SDEs that cannot be captured by previous flow-based models, while maintaining the expressivity of dynamic NFs. We corroborate these findings by numerical experiments on several well-known processes.

The remainder of this paper is organized as follows: we first provide, in section \ref{background}, an overview of the neural SDEs, wherein both the drift and diffusion are neural networks, and the dynamic normalizing flow approach and a discussion on the inherent limitations of such models. Afterward, in section~\ref{tcnf}, we introduce our model and describe its characteristics and the training algorithm. Finally, quantitative results are presented in section~\ref{quantResults} and compared to other flow-based models, and provide concluding remarks in section~\ref{conclusion}.

\begin{figure}[ht]
\begin{center}
\includegraphics[width=0.9\linewidth]{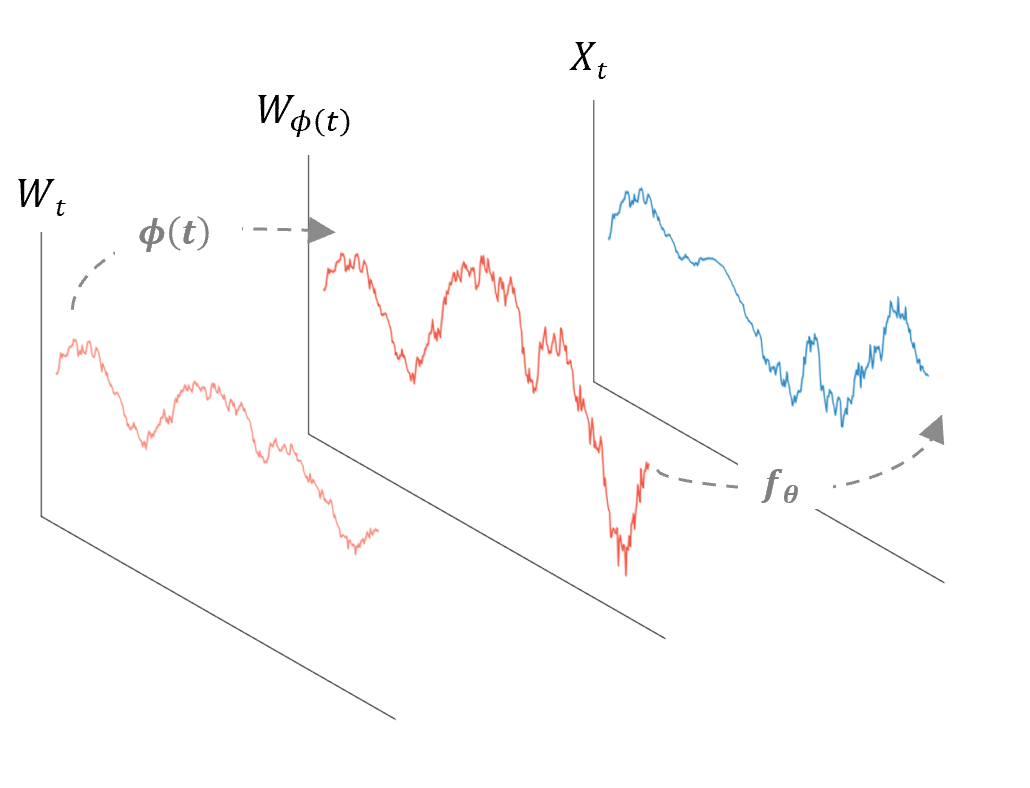}
\caption{A time-change $\phi$ is applied to the Wiener process (red) to create a new Gaussian process, which is then mapped through a bijection $f_\theta$ to the observed process (blue).}
\end{center}
\vspace{-0.75cm}
\end{figure}

\vspace{-0.2cm}
\section{Background}\label{background}
\subsection{Neural Stochastic Differential Equations}
We consider a filtered probability space $(\Omega, \mathcal{F}, P)$ and a time horizon $T$. A diffusion process $X = \{X_t\}_{t\in [0, T]}$ is defined by the Itô Stochastic Differential Equation (SDE):
\begin{align}
\label{eq:EDS_X}
    dX_t = \mu(X_t, t)dt + \sigma(X_t,t)dW_t, t\in [0, T]
\end{align}
where $W = \{W_t\}_{t\in [0, T]}$ is the $m$-dimensional adapted standard Wiener process (or Brownian motion). Functions $\mu:\mathbb{R} ^d \times [0,T] \longrightarrow \mathbb{R} ^d$ and $\sigma:\mathbb{R} ^d \times [0,T] \longrightarrow \mathbb{R} ^{d\times m}$ are the drift and diffusion coefficients, respectively. When $\mu$ and $\sigma$ are implemented by neural networks, the SDE is designated as a \textit{neural} SDE~\cite{tzen2019neural, liu2019neural}.

Several works have been proposed to learn neural SDEs using different generative modeling frameworks including Variational Autoencoders (VAEs)~\cite{li2020scalable, zeng2023latent} and Generative Adversarial Networks (GANs)~\cite{kidger2021neural}. In this paper, we specifically focus on the normalizing flow paradigm.

\vspace{-0.2cm}
\subsection{Normalizing flows}
A normalizing flow~\cite{jmlr/PapamakariosNRM21, journals/pami/KobyzevPB21, DBLP:conf/iclr/GrathwohlCBSD19} is a transformation designed to model a random variable $X$ and its corresponding complex distribution $p_X$ through a base distribution $p_Z$ and a differentiable bijective function $f:\mathbb{R} ^d  \longrightarrow \mathbb{R} ^d$. Such modeling allows for both exact density estimation and efficient sampling, by using the change of variable formula for   $X=f(Z)$: 

\begin{equation}
      \log p_X(x) = \log p_Z(z) - \log \left|\det J_{f}(z)\right| 
\end{equation}
where the Jacobian $J_f(z) = \left[\frac{\partial f_i}{\partial z_j} \right]_{1\leq i,j \leq d}$ is the $d\times d$ matrix of all partial derivatives of $f$.

Previous works have extended this framework to model time series and stochastic processes by employing a bijection that is continuously indexed by time $F(., t)$, along with a Brownian motion as the base process, giving rise to a Continuous Time Flow Process (CTFP) \cite{deng2020modeling}:
\begin{align*}
    X_t = F(W_t, t).
\end{align*}

Another approach proposed by \cite{deng2021continuous} involves incorporating latent dynamics from an Ornstein-Uhlenbeck process combined with a normalizing flow to effectively model SDEs.

These models have demonstrated the effectiveness of dynamic NFs in capturing the complex behavior of various types of stochastic processes and SDEs.
However, it is important to highlight that these models have inherent limitations. One constraint arises when applying Itô's lemma to CTFP to derive the univariate Ornstein-Uhlenbeck (OU) process described by the equation:

\begin{align}
\label{eq:ou}
    dY_t = -a(Y_t - b)dt + \sigma dW_t
\end{align}

Indeed, by applying Itô's lemma to $F(W_t,t)$ we obtain:
\begin{align}
\label{eq:NF-OU}
\begin{split}
 dF(W_t,t) = & \frac{\partial F}{\partial t}(W_t,t)dt + \frac{\partial F}{\partial x}(W_t,t)dW_t  \\
 & + \frac{1}{2}\frac{\partial^2 F}{\partial x^2}(W_t,t)dt
\end{split}
\end{align}

By comparing both Eq.~\eqref{eq:ou} and Eq.~\eqref{eq:NF-OU}, we deduce that to model the OU process, we need to have $\frac{\partial F}{\partial x}(W_t,t) = \sigma $, implying that $F(W_t, t) = \sigma W_t + g(t)$, where $g$ is a given differentiable function. However, differentiating this relation wrt $t$ and plugging in~\eqref{eq:ou} leads to the following condition:
\begin{align}\label{OUabsurde}
    \frac{dg}{dt}(t) +a g(t) -a b= -a\sigma W_t
\end{align}
Eq.~\eqref{OUabsurde} is not feasible as the left-hand side is a deterministic function of $t$ while the right-hand side is a stochastic one depending on $W_t$. Thus, CTFP shows limitations and falls short in its ability to model various stochastic processes effectively.

In the following section, we propose a model that can address this limitation and achieve improved results.

\section{Dynamic Normalizing flow with time-change}\label{tcnf}
\subsection{Time-changed Normalizing Flow}
We propose to model an observed stochastic process, denoted as $X = \{X_t\}_{t\in [0, T]}$, by combining a normalizing flow and a time-changed Wiener process to accurately capture the dynamic behavior of $X_t$ based on a realized time series $\{(x_{t_i},t_i)\}_{i=1}^n$. In this paper, we explicitly address the univariate case, with ongoing development of the general case which requires appropriate time-change for each dimension. We introduce the concept of a time-changed normalizing flow (TCNF), defined as follows:

\vspace{-0.2cm}
\begin{align*}
    X_t = f_\theta \left(W_{\phi(t)},\phi(t)\right), \quad \forall t \in [0,T],
    \vspace{-0.2cm}
\end{align*}
where $f_\theta(.,t):\mathbb{R}  \longrightarrow \mathbb{R}$ is a differentiable bijection parameterized by $\theta$, while $W_{\phi(t)}$ denotes a Brownian motion with a time-change~\cite[Sect. 1 Chap. 0]{revuz2013continuous}. The time-change is given by $\phi : \mathbb{R} ^{+} \longrightarrow \mathbb{R} ^{+}$, which is a measurable, positive and increasing function. The measurable and positive properties ensure the correct definition of $W_{\phi(t)}$, while the increasing property ensures the existence of its moments. Consequently, the neural network modeling the time-change must have intrinsically positive and increasing attributes.
The property of time change has important applications as it produces a family of Gaussian processes that are more general than Brownian motion. The Dubins-Schwarz theorem \cite[Theorem 5.1.6]{revuz2013continuous} further emphasizes this property as it states that every local martingale is simply a time-changed Brownian motion.

Therefore, by making the base process of our model a time-changed Brownian motion, we can accurately capture all instances of local martingales and semimartingales, thus generalizing the CTFP setting.
In fact, the solution to Eq.~\eqref{eq:ou} can be expressed as follows: 
\vspace{-0.2cm}
\begin{align}
    Y_t = Y_0e^{-at} + b(1-e^{-at}) + \frac{\sigma e^{-at}}{\sqrt{2a}}W_{e^{2at}-1}
\end{align}
which can be properly modeled by TCNF. More general cases like processes with time-dependent volatility can also be expressed via a time-change and therefore modeled by TCNF.
Finally, for $\phi(t) = t$ we recover the CTFP setting, which is suitable for modeling SDEs that do not require a time-change as the Geometric Brownian Motion (GBM)~\cite{oksendal2013stochastic}.

\vspace{-0.4cm}
\subsection{Time-change function}
To handle the time-change function, we employ a convex neural network that ensures a positive gradient, thereby guaranteeing a monotone output. Specifically, we utilize the M-MGN architecture~\cite{chaudhari2023learning} based on $K$ network modules defined as follows:
\begin{align}
\begin{split}
    \Tilde{t}_k &= W_k\times t + b_k, \\
    \text{M-MGN}(t) &= a + V^\top V t + \sum_{k=1}^K s_k(\Tilde{t}_k) \times W_k^\top \sigma_k(\Tilde{t}_k)
\end{split}
\end{align}

where $W_k, b_k \in \mathbb{R} ^{l \times 1}$ are respectively weight and bias vectors of the $k^{th}$ layer, $\sigma_k:\mathbb{R} ^l \longrightarrow \mathbb{R} ^l$ is an activation function and $s_k :\mathbb{R} ^l \longrightarrow \mathbb{R}$ its antiderivative. $a  \in \mathbb{R}$ and $V \in \mathbb{R} ^{l \times 1}$ are additional network parameters. As the result of M-MGN is not necessarily positive, we apply a translation of the output to ensure that the time-change is positive.

\subsection{Training algorithm}
The purpose is to train the TCNF in order to maximize the log-likelihood of the observed dataset $\{(x_{t_i},t_i)\}_{i=1}^n$:
\begin{align}\label{eq:LL}
    L = \log p_{X_{t_1}, ..., X_{t_n}}(x_{t_1}, ..., x_{t_n})
\end{align}
To compute Eq.\eqref{eq:LL}, we use the change of variable formula and leverage the independence of increments $W_{\phi(t_i)} - W_{\phi(t_{i-1})}$. Thus, the log-likelihood is expressed as follows:
\begin{align}
\begin{split}
    L = \sum_{i=1}^n & \log p_{W_{\phi(t_i)}|W_{\phi(t_{i-1})}}\left(w_{\phi(t_i)}\right) \\
    &- \log \left|\det \frac{\partial f_\theta \left(w_{\phi(t_i)},\phi(t_i)\right)}{\partial W_{\phi(t_i)}} \right|,
\end{split}
\end{align}
where $w_{\phi(t_i)} = f_\theta ^{-1} \left(x_{t_i}; \phi(t_i)\right)$ and $p_{W_{\phi(t_i)}|W_{\phi(t_{i-1})}}$ denotes the conditional Gaussian distribution with mean $W_{\phi(t_{i-1})}$ and variance $\phi(t_i)- \phi(t_{i-1})$. This constitutes one notable difference from the log-likelihood of CTFP which uses a Gaussian distribution with the same mean but variance $t_i - t_{i-1}$.

\section{Experiments}\label{quantResults}
\subsection{Toy datasets}
To evaluate the performance of our proposed model, we conducted experiments on three toy datasets comprising univariate unitless time series. These datasets were generated by sampling from three different stochastic processes. Also, in our experiments we used an identical architecture as CTFP, leveraging CNFs.

The first dataset (\textbf{Toy-SDE1}) was generated by discretizing the OU process, given by the equation: $dX_t = -\theta(X_t - \mu)dt + \sigma dW_t$, where $\mu$ and $\sigma$ represent constant parameters for the drift and volatility terms, respectively. Parameter $\theta$ captures the speed at which a given sample path of the process converges towards the drift term. This dataset aims to assess the model's ability to capture the dynamics of time changes.

The second dataset (\textbf{Toy-SDE2}) was generated based on the equation: $dX_t = -\theta(X_t - \mu)dt + \sigma \sqrt{t}dW_t$, describing an OU with a time-dependent diffusion coefficient, and is used to test the model's capability to capture time transformations with increasing complexity. Notably, this SDE is of interest as it is commonly used in score-based models \cite{yang2022diffusion}, where noise is gradually introduced during the training process.

Finally, The third dataset (\textbf{Toy-SDE3}) involved the geometric Brownian motion, described by the equation: $dX_t = \mu X_t dt + \sigma X_t dW_t$, where $\mu$ and $\sigma$ represent constant parameters for the drift and volatility terms, respectively. 
This dataset was designed to showcase the capacity of TCNF to handle SDEs where no time-change is required, effectively learning the simple function $\phi(t) = t$. This shows that our approach can encompass the CTFP framework.

\begin{table*}[ht]
\centering
\begin{tabular}{llcccc}
\toprule
Dataset & Model & $m_{X_t}$ & $\sigma_{X_t}$ & IQR &$p_{X_t}$ \\
\hline
\midrule
\multirow{2}{*}{Toy-SDE1} & CTFP & $0.022 \pm 0.009$ & $0.241 \pm 0.012$ & $0.886 \pm 0.021$ & $0.062$ \\
                          & TCNF & $\mathbf{0.019} \pm \mathbf{0.007}$ & $\mathbf{0.109} \pm \mathbf{0.009}$ & $\mathbf{0.700} \pm \mathbf{0.014}$ & $\mathbf{0.033}$ \\

\midrule
\multirow{2}{*}{Toy-SDE2} & CTFP & $0.059 \pm 0.011$ & $0.523 \pm 0.018$ & $1.517 \pm 0.037$ & $0.087$ \\
                          & TCNF & $\mathbf{0.024} \pm \mathbf{0.012}$ & $\mathbf{0.322} \pm \mathbf{0.016}$ & $\mathbf{1.183} \pm \mathbf{0.024}$ & $\mathbf{0.061}$ \\
\midrule
\multirow{2}{*}{Toy-SDE3} & CTFP & $\mathbf{0.035} \pm \mathbf{0.020}$ & $\mathbf{0.086} \pm \mathbf{0.056}$ & $1.103 \pm 0.037$ & $\mathbf{0.005}$ \\
                          & TCNF & $0.036 \pm 0.018$ & $0.099 \pm 0.072$ & $\mathbf{1.083} \pm \mathbf{0.038}$ & $0.007$ \\
\bottomrule
\end{tabular}
\caption{Quantitative analysis: We display mean absolute errors (MAE) for estimating respectively the mean, standard deviation, interquartile range IQR = $Q_3 - Q_1$, and density. The estimations are based on 1000 sample paths over 1000 iterations.}
\label{tab:quant_error_toy12}
\end{table*}

\begin{figure}[ht]
\begin{center}
    \includegraphics[width=0.925\linewidth]{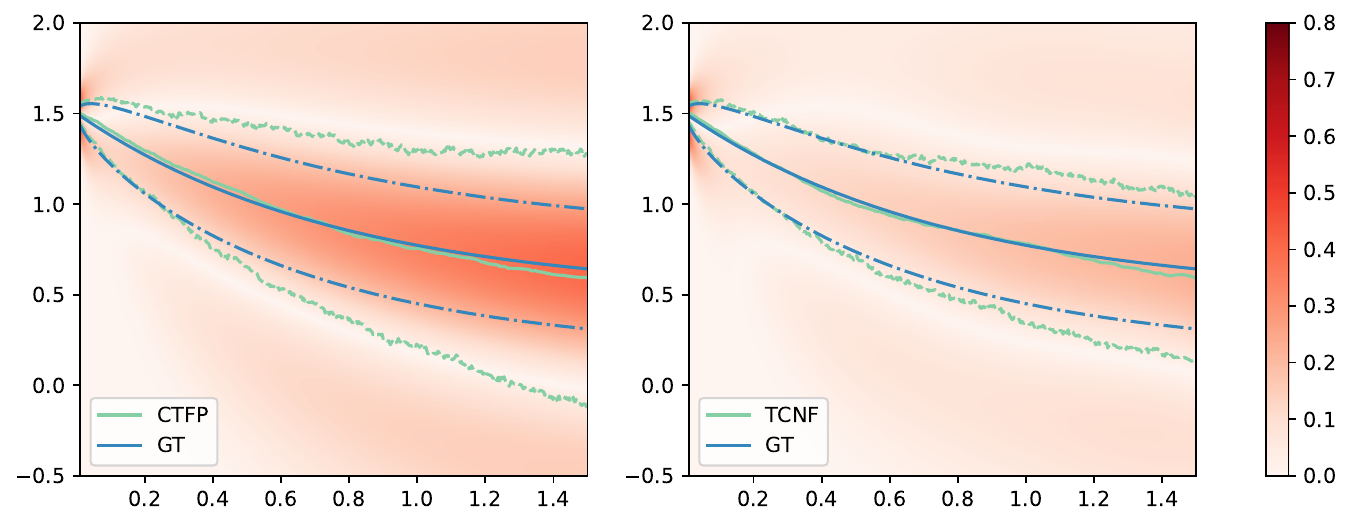}
    \includegraphics[width=0.925\linewidth]{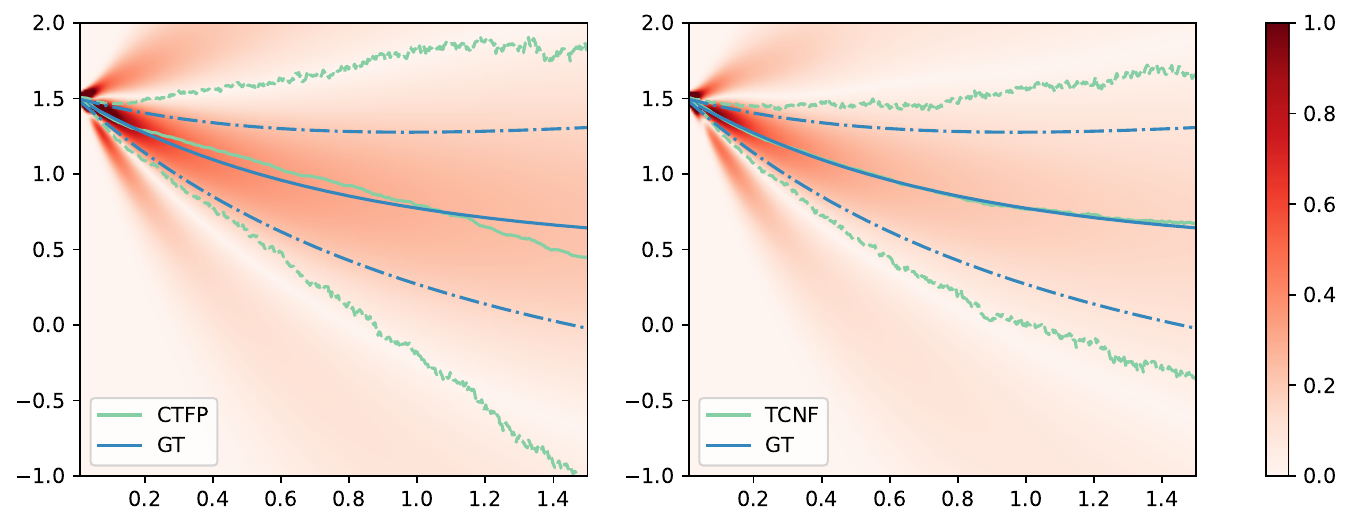}   
\caption{Comparison between TCNF and CTFP on toy-SDE1 (top) and toy-SDE2 (bottom). In each figure we depict density MAE between the flow models and the ground truth (GT), along the mean (continuous line) and the IQR (dashed line).}
\label{fig:result_density_toy12}
\end{center}
\end{figure}

The quantitative comparison is carried by comparing estimations of the mean $m_{X_t}$, the standard-deviation $\sigma_{X_t}$, the interquartile range IQR$=Q_{3} - Q_{1}$, and the density $p_{X_t}$ respectively. For each model, we compute the mean absolute errors (MAE) against the ground-truth values. The mean, standard deviation, and quartiles are estimated based on 1000 sample paths over 1000 iterations, while the density is estimated by the change of variable formula on a grid consisting of 1000 spatial points and 500 temporal points within the time interval $[0, T = 1.5]$. The quantitative results reported in Table \ref{tab:quant_error_toy12} and Figure \ref{fig:result_density_toy12} show first that TCNF shows no loss in generality as it handles cases where no time-change is required, and second that our model exhibits superior estimation capability, as it can capture the behavior of time-changed ground truth solutions.

\vspace{-0.2cm}
\subsection{Real-world datasets}
To further assess our model's capacity to capture complex dynamics, we train it on two real-world datasets: Crypto-forecasting (\textbf{Crypto})~\cite{g-research-crypto-forecasting} and Electric Consumption Load (\textbf{ECL})~\cite{zhou2021informer}. The Crypto dataset contains historical prices of various cryptocurrencies. We focused our analysis on modeling Ethereum log returns over the 2020 period.
The ECL dataset comprises electricity consumption data from multiple clients within a 15-minute interval. We choose to model the consumption of client '200' for its extended time series.

The results comprise mean absolute errors (MAE) for estimating the mean ($m_{X_t}$) and standard deviation ($\sigma_{X_t}$) for the Crypto dataset. We employed mean relative errors (MRE) for the ECL dataset to scale the results appropriately. These results are reported in Table \ref{tab:real-world} and compared against CTFP.

\begin{table}
\centering
\begin{tabular}{llcc}
\toprule
Dataset & Model & $m_{X_t}$ & $\sigma_{X_t}$ \\
\hline
\midrule
\multirow{2}{*}{Crypto} & CTFP & $0.083 \pm 0.007$ & $0.232 \pm 0.009$\\
                        & TCNF & $\mathbf{0.025} \pm \mathbf{0.001}$ & $\mathbf{0.131} \pm \mathbf{0.005}$\\
\midrule

\multirow{2}{*}{ECL} & CTFP & $0.771 \pm 0.108$ & $26.972 \pm 8.634$ \\
                     & TCNF & $\mathbf{0.299} \pm \mathbf{0.007}$ &  $\mathbf{2.182} \pm \mathbf{2.081}$ \\
\bottomrule
\end{tabular}
\caption{Quantitative analysis of real-world dataset: We display mean and standard deviation estimation errors.}
\label{tab:real-world}
\end{table}

\vspace{-0.2cm}
\section{Conclusion}\label{conclusion}
We introduced a generalized approach to modeling SDEs via dynamic NFs and time-change. By transforming the Wiener process through time, we generate various Gaussian processes, which are then mapped to the observed process via a bijection. The time-change combined with the NF enabled us to model processes that are otherwise challenging to derive due to calculus constraints. Importantly, this extension retains the advantages of dynamic NFs, such as exact density estimation and efficient sampling.

Experiments showed that our model exhibits better performance and a generalization capacity. We believe incorporating dimension-specific time-changes enables us to extend the method to higher dimensions. Additionally, improvement in the calibration of the time-change can be achieved by linking it to either the quadratic variation of the process or its moments.

\bibliographystyle{IEEEbib}
\bibliography{references}

\begin{thebibliography}{10}

\bibitem{oksendal2013stochastic}
Bernt Oksendal,
\newblock {\em Stochastic differential equations: an introduction with applications},
\newblock Springer Science \& Business Media, 2013.

\bibitem{conf/nips/GoodfellowPMXWOCB14}
Ian~J. Goodfellow, Jean Pouget{-}Abadie, Mehdi Mirza, Bing Xu, David Warde{-}Farley, Sherjil Ozair, Aaron~C. Courville, and Yoshua Bengio,
\newblock ``Generative adversarial nets,''
\newblock in {\em Advances in Neural Information Processing Systems 27}, 2014.

\bibitem{journals/corr/KingmaW13}
Diederik~P. Kingma and Max Welling,
\newblock ``Auto-encoding variational bayes,''
\newblock in {\em International Conference on Learning Representations}, 2014.

\bibitem{jmlr/PapamakariosNRM21}
George Papamakarios, Eric~T. Nalisnick, Danilo~Jimenez Rezende, Shakir Mohamed, and Balaji Lakshminarayanan,
\newblock ``Normalizing flows for probabilistic modeling and inference,''
\newblock {\em J. Mach. Learn. Res.}, vol. 22, 2021.

\bibitem{journals/pami/KobyzevPB21}
Ivan Kobyzev, Simon J.~D. Prince, and Marcus~A. Brubaker,
\newblock ``Normalizing flows: An introduction and review of current methods,''
\newblock {\em {IEEE} Trans. Pattern Anal. Mach. Intell.}, vol. 43, 2021.

\bibitem{conf/nips/SongE19}
Yang Song and Stefano Ermon,
\newblock ``Generative modeling by estimating gradients of the data distribution,''
\newblock in {\em Advances in Neural Information Processing Systems 32}, 2019.

\bibitem{conf/nips/YoonJS19}
Jinsung Yoon, Daniel Jarrett, and Mihaela van~der Schaar,
\newblock ``Time-series generative adversarial networks,''
\newblock in {\em Advances in Neural Information Processing Systems 32}, 2019.

\bibitem{kidger2021neural}
Patrick Kidger, James Foster, Xuechen Li, and Terry~J Lyons,
\newblock ``Neural sdes as infinite-dimensional gans,''
\newblock in {\em International conference on machine learning}. PMLR, 2021.

\bibitem{li2020scalable}
Xuechen Li, Ting-Kam~Leonard Wong, Ricky~TQ Chen, and David~K Duvenaud,
\newblock ``Scalable gradients and variational inference for stochastic differential equations,''
\newblock in {\em Symposium on Advances in Approximate Bayesian Inference}. PMLR, 2020.

\bibitem{zeng2023latent}
Sebastian Zeng, Florian Graf, and Roland Kwitt,
\newblock ``Latent sdes on homogeneous spaces,''
\newblock {\em arXiv preprint arXiv:2306.16248}, 2023.

\bibitem{mehrasa2019point}
Nazanin Mehrasa, Ruizhi Deng, Mohamed~Osama Ahmed, Bo~Chang, Jiawei He, Thibaut Durand, Marcus Brubaker, and Greg Mori,
\newblock ``Point process flows,''
\newblock {\em arXiv preprint arXiv:1910.08281}, 2019.

\bibitem{conf/iclr/ShchurBG20}
Oleksandr Shchur, Marin Bilos, and Stephan G{\"{u}}nnemann,
\newblock ``Intensity-free learning of temporal point processes,''
\newblock in {\em International Conference on Learning Representations}, 2020.

\bibitem{DBLP:conf/iclr/GrathwohlCBSD19}
Will Grathwohl, Ricky T.~Q. Chen, Jesse Bettencourt, Ilya Sutskever, and David Duvenaud,
\newblock ``{FFJORD:} free-form continuous dynamics for scalable reversible generative models,''
\newblock in {\em International Conference on Learning Representations}, 2019.

\bibitem{deng2020modeling}
Ruizhi Deng, Bo~Chang, Marcus~A Brubaker, Greg Mori, and Andreas Lehrmann,
\newblock ``Modeling continuous stochastic processes with dynamic normalizing flows,''
\newblock {\em Advances in Neural Information Processing Systems 33}, 2020.

\bibitem{deng2021continuous}
Ruizhi Deng, Marcus~A Brubaker, Greg Mori, and Andreas Lehrmann,
\newblock ``Continuous latent process flows,''
\newblock {\em Advances in Neural Information Processing Systems 34}, 2021.

\bibitem{tzen2019neural}
Belinda Tzen and Maxim Raginsky,
\newblock ``Neural stochastic differential equations: Deep latent gaussian models in the diffusion limit,''
\newblock {\em arXiv preprint arXiv:1905.09883}, 2019.

\bibitem{liu2019neural}
Xuanqing Liu, Tesi Xiao, Si~Si, Qin Cao, Sanjiv Kumar, and Cho-Jui Hsieh,
\newblock ``Neural sde: Stabilizing neural ode networks with stochastic noise,''
\newblock {\em arXiv:1906.02355}, 2019.

\bibitem{revuz2013continuous}
Daniel Revuz and Marc Yor,
\newblock {\em Continuous martingales and Brownian motion}, vol. 293,
\newblock Springer Science \& Business Media, 2013.

\bibitem{chaudhari2023learning}
Shreyas Chaudhari, Srinivasa Pranav, and Jos{\'e}~MF Moura,
\newblock ``Learning gradients of convex functions with monotone gradient networks,''
\newblock in {\em ICASSP 2023-2023 IEEE International Conference on Acoustics, Speech and Signal Processing (ICASSP)}. IEEE, 2023, pp. 1--5.

\bibitem{yang2022diffusion}
Ling Yang, Zhilong Zhang, Yang Song, Shenda Hong, Runsheng Xu, Yue Zhao, Yingxia Shao, Wentao Zhang, Bin Cui, and Ming-Hsuan Yang,
\newblock ``Diffusion models: A comprehensive survey of methods and applications,''
\newblock {\em arXiv preprint arXiv:2209.00796}, 2022.

\bibitem{g-research-crypto-forecasting}
Alessandro Ticchi, Andrew Scherer, Carla McIntyre, Carlos Stein~N Brito, Derek Snow, Develra, dstern, James Colless, Kieran Garvey, Maggie, Maria~Perez Ortiz, Ryan Lynch, and Sohier Dane,
\newblock ``G-research crypto forecasting,'' \url{https://kaggle.com/competitions/g-research-crypto-forecasting}, 2021,
\newblock Kaggle.

\bibitem{zhou2021informer}
Haoyi Zhou, Shanghang Zhang, Jieqi Peng, Shuai Zhang, Jianxin Li, Hui Xiong, and Wancai Zhang,
\newblock ``Informer: Beyond efficient transformer for long sequence time-series forecasting,''
\newblock in {\em Proceedings of the AAAI conference on artificial intelligence}, 2021, vol.~35, pp. 11106--11115.

\end{thebibliography}
\end{document}